%% file: main.tex
\documentclass[10pt,twocolumn,letterpaper]{article}

\usepackage{mystyle}   
\usepackage{graphicx}
\usepackage{amsmath}
\usepackage{amssymb}
\usepackage{booktabs}
\usepackage[noend]{algorithm2e}
\usepackage{setspace}
\usepackage{float}
\usepackage{multirow}
\usepackage{array}

\usepackage[pagebackref,breaklinks,colorlinks]{hyperref}

\usepackage[capitalize]{cleveref}
\crefname{section}{Sec.}{Secs.}
\Crefname{section}{Section}{Sections}
\Crefname{table}{Table}{Tables}
\crefname{table}{Tab.}{Tabs.}

\input{99_preamble}
\begin{document}
\title{\moniker: Hierarchical Graphs for Generalized Modelling of Clothing Dynamics}

\author{Artur Grigorev$^{1,2}$
\and
Bernhard Thomaszewski$^{1}$
\and
Michael J. Black$^{2}$\\
\and
Otmar Hilliges$^{1}$
\\
\\ $^1$ ETH Zurich, Department of Computer Science 
\\ $^2$ Max Planck Institute for Intelligent Systems, Tubingen 
\\\\ \href{https://dolorousrtur.github.io/hood/}{https://dolorousrtur.github.io/HOOD/}
}

\twocolumn[{
\maketitle
\begin{center}
    \captionsetup{type=figure}
    \includegraphics[width=1.\textwidth]{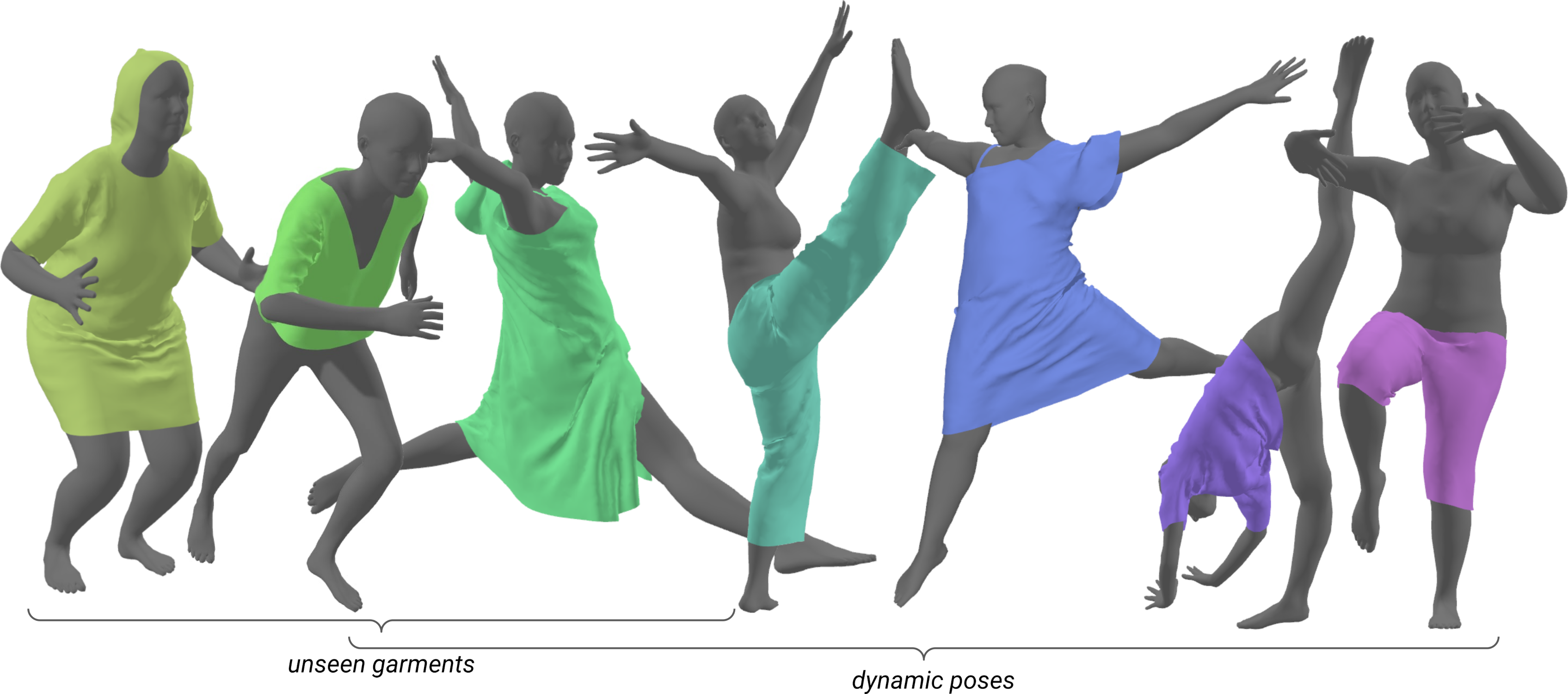}
    \captionof{figure}{
    We combine graph neural networks, a hierarchical graph representation and multi-level message passing with an unsupervised training scheme to enable real-time prediction of realistic clothing dynamics for arbitrary types of garments and body shapes.
    Our method models both tight-fitting and free-flowing clothes draped over arbitrary body shapes. The method generalizes to new, entirely unseen, garments (left), and allows for dynamic and unconstrained poses (right) and changes in material parameters and topology at test time.  
    }
    \label{fig:teaser}
\end{center}
}]
\input{00_abstract}

\input{01_introduction}
\input{02_related}
\input{03_method2}

\input{04_experiments}
\input{05_limitations}

{\small
\bibliographystyle{ieee_fullname}
\bibliography{egbib}
}

\end{document}

%% file: 99_preamble.tex
\usepackage{overpic}
\usepackage{enumitem} %
\usepackage{overpic} %
\usepackage{color}
\usepackage[dvipsnames]{xcolor}
\usepackage{paralist}
\usepackage[utf8]{inputenc}
\usepackage{soul} %
\usepackage{graphicx}
\usepackage[export]{adjustbox}

\definecolor{turquoise}{cmyk}{0.65,0,0.1,0.3}
\definecolor{purple}{rgb}{0.65,0,0.65}
\definecolor{dark_green}{rgb}{0, 0.5, 0}
\definecolor{orange}{rgb}{0.8, 0.6, 0.2}
\definecolor{red}{rgb}{0.8, 0.2, 0.2}
\definecolor{darkred}{rgb}{0.6, 0.1, 0.05}
\definecolor{blueish}{rgb}{0.0, 0.3, .6}
\definecolor{light_gray}{rgb}{0.7, 0.7, .7}
\definecolor{pink}{rgb}{1, 0, 1}
\definecolor{greyblue}{rgb}{0.25, 0.25, 1}

\newif\ifshowcomments
\showcommentstrue %
\ifshowcomments

\newcommand{\moniker}{HOOD\xspace}

%% file: 00_abstract.tex
\begin{abstract}
\vspace{-7mm}
We propose a method that leverages graph neural networks, multi-level message passing, and unsupervised training to enable efficient prediction of realistic clothing dynamics.
Whereas existing methods based on linear blend skinning must be trained for specific garments, our method, called HOOD, is agnostic to body shape and applies to tight-fitting garments as well as loose, free-flowing clothing.
Furthermore, HOOD handles changes in topology (e.g., garments with buttons or zippers) and material properties at inference time. 
As one key contribution, we propose a hierarchical message-passing scheme that efficiently propagates stiff stretching modes while preserving local detail. We empirically show that HOOD outperforms strong baselines quantitatively and that its results are perceived as more realistic than state-of-the-art methods. 

\end{abstract}

%% file: 01_introduction.tex
\section{Introduction}
\label{sect:intro}

The ability to model realistic and compelling clothing behavior is crucial for telepresence, virtual try-on, video games, and many other applications that rely on high-fidelity digital humans. A common approach to generating plausible dynamic motions is physics-based simulation~\cite{Baraff98Large}.
While impressive results can be obtained, physical simulation is sensitive to initial conditions, requires animator expertise, and is computationally expensive; state-of-the-art approaches \cite{Narain12Adaptive,Cirio14Yarnlevel,Li21Codimensional} are not designed for the strict computation budgets imposed by real-time applications.

Deep learning-based methods have started to show promising results both in terms of efficiency and quality.
However, there are several limitations that have so far prevented such approaches from unlocking their full potential:

First, existing methods rely on linear-blend skinning and compute clothing deformations primarily as a function of body pose~\cite{santesteban2022snug,li2022dig}. While compelling results can be obtained for tight-fitting garments such as shirts and sportswear, skinning-based methods struggle with dresses, skirts, and other types of loose-fitting clothing that do not closely follow body motion. 

Crucially, many state-of-the-art learning-based methods are garment-specific~\cite{Santesteban_2021_CVPR,santesteban2022snug,zhang2022motion,pan2022predicting,guan2012drape} and can only predict deformations for the particular outfit they were trained on. 
The need to retrain these methods for each garment limits applicability.

In this paper, we propose a novel approach for predicting dynamic garment deformations using graph neural networks (GNNs).
Our method learns to predict physically-realistic fabric behavior by reasoning about the map between local deformations, forces, and accelerations. Thanks to its locality, our method is agnostic to both the global structure and shape of the garment and directly generalizes to arbitrary body shapes and motions. 

GNNs have shown promise in replacing physics-based simulation \cite{pfaff2020learning,sanchez2020learning}, but a straightforward application of this concept to clothing simulation yields unsatisfying results.

GNNs apply local transformations (implemented as MLPs) to feature vectors of vertices and their one-ring neighborhood in a given mesh. 
Each transformation results in a set of messages that are then used to update feature vectors. 
This process is repeated, allowing signals to propagate through the mesh. 
However, a fixed number of message-passing steps limits signal propagation to a finite radius. 
This is problematic for garment simulation, where elastic waves due to stretching travel rapidly through the material, leading to quasi-global and immediate long-range coupling between vertices.
Using too few steps delays signal propagation and leads to disturbing over-stretching artifacts, making garments look unnatural and rubbery. 
Na\"{i}vely increasing the number of iterations comes at the expense of rapidly growing computation times. 
This problem is amplified by the fact that the maximum size and resolution of simulation meshes is not known \textit{a priori}, which would allow setting a conservative, sufficiently large number of iterations.

To address this problem, we propose a message-passing scheme over a hierarchical graph that interleaves propagation steps at different levels of resolution. In this way, fast-travelling waves due to stiff stretching modes can be efficiently treated on coarse scales, while finer levels provide the resolution needed to model local detail such as folds and wrinkles. We show through experiments that our graph representation improves predictions both qualitatively and quantitatively for equal computation budgets.

To extend the generalization capabilities of our approach, we combine the concepts of graph-based neural networks and differentiable simulation by using an incremental potential for implicit time stepping as a loss function~\cite{Martin11ExampleBased,santesteban2022snug}. 

This formulation allows our network to be trained in a fully unsupervised way and to simultaneously learn multi-scale clothing dynamics, the influence of material parameters, as well as collision reaction and frictional contact with the underlying body, without the need for any ground-truth (GT) annotations.
Additionally, the graph formulation enables us to model garments of varied and {\em changing} topology; e.g.~the unbuttoning of a shirt in motion.

In summary, we propose a method, called HOOD, that leverages graph neural networks, multi-level message passing, and unsupervised training to enable 
real-time prediction of realistic clothing dynamics for arbitrary types of garments and body shapes. 

We empirically show that our method offers strategic advantages in terms of flexibility and generality compared to state-of-the-art approaches. Specifically, we show that a single trained network:
(i) efficiently predicts physically-realistic dynamic motion for a large variety of garments;
(ii) generalizes to new garment types and shapes not seen during training;
(iii) allows for run-time changes in material properties and garment sizes, and
(iv) supports dynamic topology changes such as opening zippers or unbuttoning shirts.
Code and models are available for research purposes: \url{https://dolorousrtur.github.io/hood/}.

%% file: 02_related.tex
\section{Related Work}
\label{sect:related}

\textbf{Physics-based Simulation.}
Modeling the behavior of 3D clothing is a longstanding problem in computer graphics \cite{Terzopoulos87Elastically}. 
Central research problems include mechanical modeling \cite{choi2005stable,Grinspun03Discrete,Volino09ASimple}, 
material behavior \cite{bhat2003estimating,Wang11DataDriven,Miguel12DataDriven}, time integration \cite{baraff1998large,Thomaszewski2008AsynchronousCS}, collision handling \cite{bridson2002robust,Harmon09ACM,tang2018cloth,Li21Codimensional} and, more recently, differentiable simulation \cite{Liang19Differentiable,Li22DiffCloth}.
While state-of-the-art methods can generate highly realistic results with impressive levels of detail, physics-based simulation methods are often computationally expensive.

\textbf{Learned Deformation Models.}
To overcome the performance limitations of traditional physical simulators, prior work uses machine learning to accelerate computation. One line of research uses neural networks, in combination with linear blend skinning, to learn garment deformations from pose and shape parameters of human body models~\cite{guan2012drape, santesteban2019learning, ma2020learning, patel2020tailornet}. While such methods can produce plausible results at impressive rates, their reliance on skinning limits their ability to realistically model loose, free-flowing garments such as long skirts and dresses.

Several learning-based methods specifically tackle loose garments. For example, Santesteban et al.~\cite{Santesteban_2021_CVPR} introduce a diffused body model to extend blend shapes and skinning weights to any 3D point beyond the body mesh. Pan et al.~\cite{pan2022predicting} use virtual bones to drive the shape of garments as a function of the body pose. 

A common limitation of such methods is their inability to generalize across multiple garments: as they predict a specific number of vertex offsets, these networks need to be retrained even for small changes in garment geometry.

Another set of methods learns  geometric priors from a large dataset of synthetic garments~\cite{bertiche2020cloth3d}. Zakharkin et al.~\cite{zakharkin2021point} learn a latent space of garment geometries, modeling garments as point clouds.  

Su et al.~\cite{su2022deepcloth} represent garments in texture space, allowing them to explicitly control garment shape and topology. DeePSD \cite{bertiche2021deepsd} learns a mapping from a garment template to its skinning weights. Although these methods are able to generate geometries for various garments, including loose ones, they still rely on linear blend skinning and parametric body models, ultimately limiting their ability to generate dynamic garment motions.

Traditional, mesh-based methods are typically restricted to fixed topologies and cannot deal with topological changes such as unzipping a jacket. To address this, several methods resort to implicit shape models \cite{Aggarwal:ACCV:2020,POP:ICCV:2021,santesteban2021ulnefs,Buffet:2019,corona2021smplicit,Chen:ICCV:2021,lin2022fite}.
While these can capture arbitrary topology, they require significant training data, are not compatible with existing graphics pipelines, and are expensive to render.
In contrast, our graphical formulation supports varied topologies with an efficient mesh representation.

Recent work learns the mapping from body parameters to garment deformations in an unsupervised way. PBNS \cite{bertiche2020pbns} pioneered this idea by using the potential energy of a mass-spring system to train a neural network to predict garment deformations in static equilibrium. 
SNUG~\cite{santesteban2022snug} extends this approach with a dynamic component, using a recurrent neural network to predict sequences of garment deformations. Their method outperforms other learning-based approaches without using any physically-simulated training data. SNUG, however, is limited to tight-fitting garments and cannot generalize to novel garments.

Our method builds on the idea of a physics-based loss function for self-supervised learning. Unlike SNUG, however, our method does not rely on skinning and is able to generate plausible dynamic motion for arbitrary types of garments, including dynamic, free-flowing clothing and changing topology.

\textbf{Graph-based Methods.}
As another promising line of research, graph neural networks have recently started to show their potential for learning-based modeling of dynamic physical systems \cite{sanchez2020learning,pfaff2020learning,fortunato2022multiscale}. Most closely related to our work are MeshGraphNets~\cite{pfaff2020learning}, which use a message-passing network to learn the dynamics of physical systems such as fluids and cloth from mesh-based simulations. 
Due to their local nature, MeshGraphNets achieve outstanding generalization capabilities.
However, using their default message passing scheme, signals tend to propagate slowly through the graph and it is difficult to set hyper-parameters (number of message passing steps) in advance such that satisfying behavior is obtained. 

This problem is particularly relevant for clothing simulation, where an insufficient number of steps can lead to excessive stretching and unnatural dynamics. We address this problem with a multi-level message passing architecture that uses a hierarchical graph to accelerate signal propagation.

Recently, several graph pooling strategies were introduced to increase the radius of message propagation including learned pooling~\cite{gao2019graph}, pooling by rasterization~\cite{lino2021simulating} and spatial proximity~\cite{fortunato2022multiscale}.
Concurrent to ours, the work of Cao et al.~\cite{cao2022bi} analyzes limitations of pooling strategies and suggests using pre-computed coarsened geometries with hand-crafted aggregation weights for inter-level transfer. We propose a simple and efficient graph coarsening strategy that allows our network to implicitly learn transitions between graph levels, thus avoiding the need for any manually designed transfer operators.

Graph-based methods have demonstrated their ability to model numerous types of physical systems, including fabric simulation. To the best of our knowledge, however, we are the first to propose a graph-based approach for modeling the dynamic garment motions of dressed virtual humans. 

%% file: 03_method2.tex
\section{Method}
\label{sect:method}

\begin{figure*}[ht]
    \centering
    \includegraphics[width=1\textwidth]{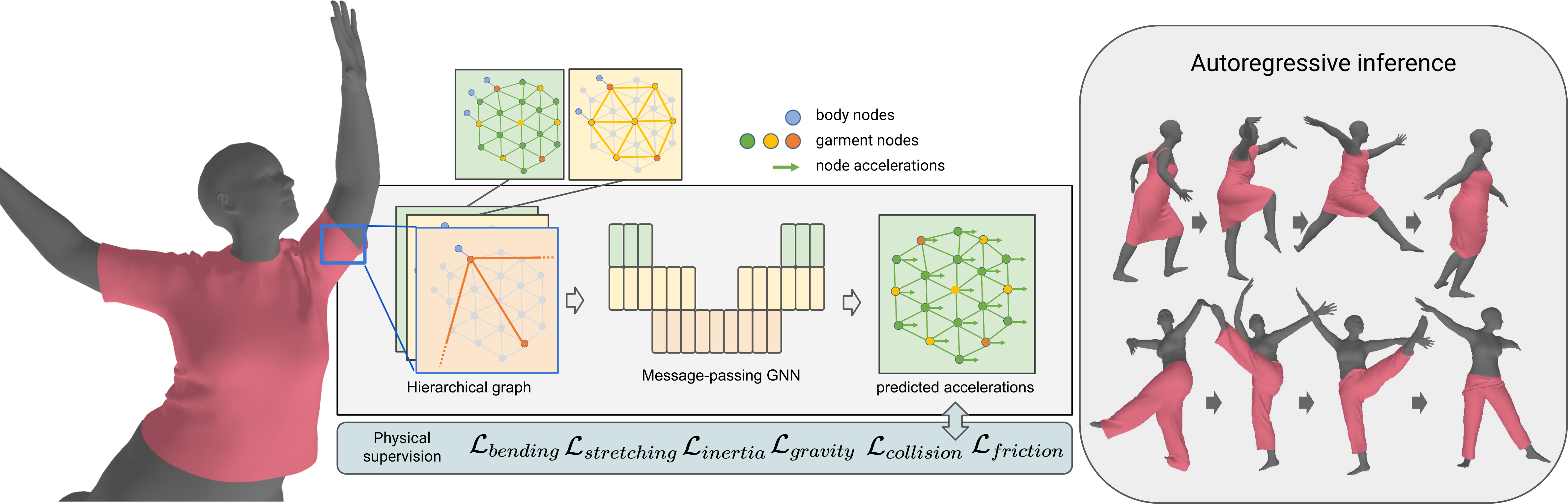}
    \caption{\textbf{Method overview}: We model garment mesh interactions based on a graph that is derived from the garment mesh and augmented with additional body nodes (blue) and edges between the garment and the closest body nodes. The input graph is converted into a hierarchical graph structure to allow  fast signal propagation and is processed by a message-passing network with a UNet-like architecture. The colors of garment nodes show which of the nodes are present on each of the hierarchical levels and correspond to the levels of the GNN. The GNN  predicts accelerations (green) for each garment node. The model is trained via self-supervision with a set of physical objectives, thus removing the need for any offline training data. At inference time, the model autoregressively generates dynamic garment motions over long time horizons. A single network can generalize to unseen garments, material properties, and even topologies.}
    \label{fig:method}
\end{figure*}

\begin{figure*}[ht]
  \centerline{  \includegraphics[width=.9\linewidth]{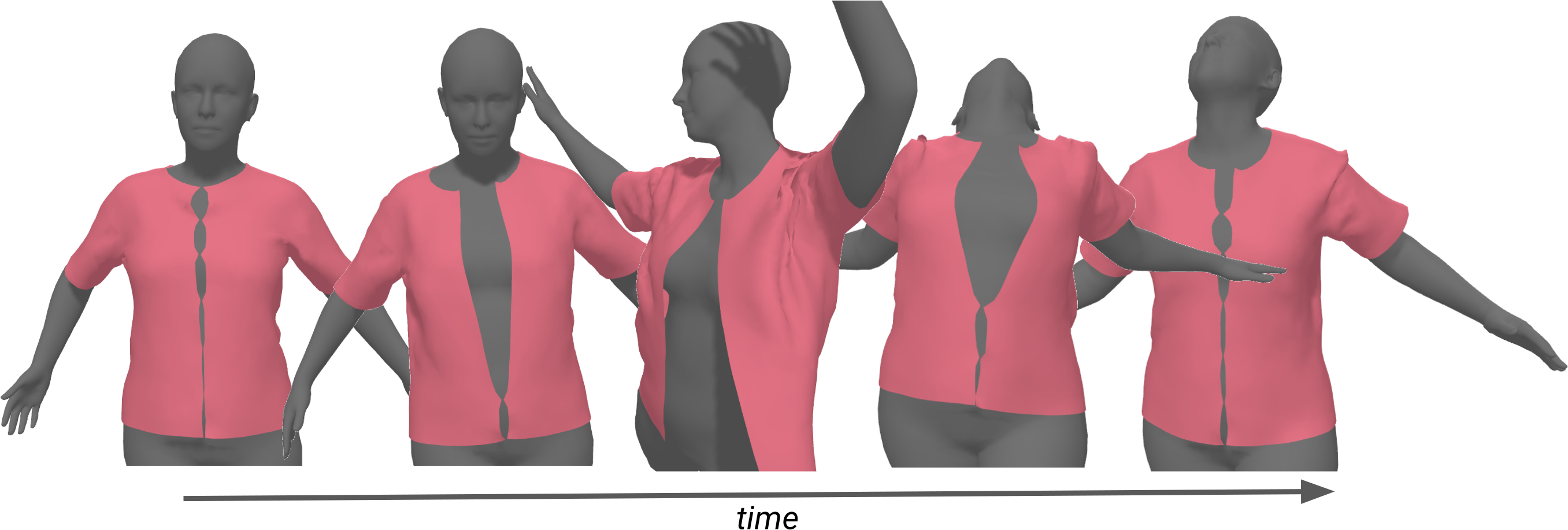}}
\vspace{-0.1in}
  \caption{Its graph-based nature allows HOOD to model garments with changing topology by enabling and disabling specific edges in the garment mesh. Here we show frames from a sequence generated with our model, where we unbutton a shirt and then button it back up.}
  \label{fig:unzip}
\end{figure*}

Our method, schematically summarized in \cref{fig:method}, learns the parameters of a single network that is able to predict plausible dynamic motion for a wide range of garment types and shapes, that generalizes to new, unseen clothing, and allows for dynamic changes in material parameters (\cref{fig:bcoef}) and garment topology (\cref{fig:unzip}).
These unique capabilities derive from a novel combination of \textit{a}) graph neural networks that learn the local dynamics in a garment-independent way (Sec. \ref{subsec:graphnet}), \textit{b}) hierarchical message-passing for the efficient capture of long-range coupling (Sec. \ref{subsec:hierarchical}), and \textit{c}) a physics-based loss function that enables self-supervised training (Sec. \ref{subsec:garmentmodel}).

\subsection{Background}
\label{subsec:graphnet}
HOOD builds on MeshGraphNets~\cite{pfaff2020learning}, a type of
graph neural network, to learn the local dynamics of deformable materials.
Once trained, MeshGraphNets predict nodal accelerations from current positions and velocities, which are then used to step the garment mesh forward in time. 

\textbf{Basic Structure.} We model garment dynamics based on a graph consisting of the vertices and edges of the garment mesh, augmented with so-called \textit{body edges}: for each garment node we find the nearest vertex on the body model and add a new edge if the distance is below a threshold $r$. Vertices and edges are endowed with feature vectors $v_i$ and $e_{ij}$, respectively, where $i$ and $j$ are node indices.
Nodal feature vectors consist of a type value (garment or body), current state variables (velocity, normal vector), and physical properties (mass). Edge feature vectors store the relative position between their two nodes w.r.t.~both the current state and canonical geometry of the garment (\cref{fig:method}, left).

\textbf{Message Passing.}
To evolve the system forward in time, we apply $N$ message-passing steps on the input graph. In each step, edge features are first updated as
\begin{equation}
e_{ij}\leftarrow f_{v\rightarrow e}(e_{ij},v_i,v_j) \ ,
\end{equation}
where $f_{v\rightarrow e}$ is a multilayer perceptron (MLP). 
Nodes are then updated by processing the average of all incident edge features, denoted by $f_{e\rightarrow v}$:
\begin{equation}
v_{i}\leftarrow f_{e\rightarrow v}(v_i, \sum_j e^{body}_{ij}, \sum_j e_{ij}) \ ,
\end{equation}
where $f_{e\rightarrow v}$ is another multilayer perceptron (MLP), and $ e^{body}_{ij}$ are body edges. While the same MLP is used for all nodes, each set of edges is processed by a separate MLP. 
After $N$ message-passing steps, the nodal features are passed into a decoder MLP to obtain per-vertex accelerations, which are then used to compute end-of-step velocities and positions. 

\textbf{Extensions for Clothing.}
To model different types of fabric and multi-fabric garments, we extend node and edge feature vectors with local material parameters. These material parameters include: Young's modulus and Poisson's ratio (mapped to their corresponding Lam{\'e} parameters $\mu$ and $\lambda$) that model the stretch resistance and area preservation of a given fabric, the bending coefficient $k_{bending}$ that penalizes folding and wrinkling, as well as the density of the fabric, defining its weight.
Since our network supports heterogeneous material properties (for each edge and node) as input, the definition of individual material parameters for different parts to model multi-material garments is possible, even at inference time (See~\cref{fig:bcoef}).

\subsection{Hierarchical Message Passing}
\label{subsec:hierarchical}
Fabric materials are sufficiently stiff such that forces applied in one location propagate 
rapidly across the garment. 
When using a fixed number of message-passing steps, however, forces can only propagate within a finite radius for a given time step. Consequently, using too small a number of message-passing steps will make garments appear overly elastic and rubbery.
We solve this problem by extending MeshGraphNets to accelerate signal propagation. To this end, we construct a hierarchical graph representation from the flat input graph and use this 
to accelerate signal propagation during message-passing.

\textbf{Hierarchical Graph Construction.} 
Allowing messages to travel further within a single step requires long-range connections between nodes. To this end, we recursively coarsen a given input graph to obtain a hierarchical representation. 
Although there are many other options for generating graph hierarchies, we take inspiration from concurrent work \cite{cao2022bi} and use a simple but effective recursive process. 
We start by partitioning the nodes of the input graph into successively coarser sets such that the inter-vertex distance --- the number of edges in the shortest path between two nodes --- increases. We then create new sets of coarsened edges for each of the partitions. See Fig.~\ref{fig:method} for an illustration and the supplemental material for details.

By applying this algorithm recursively, we obtain a \textit{nested} hierarchical graph in which the nodes of each coarser level form a proper subset of the next finer level nodes, i.e., $V_{l+1}\subset V_{l}$. This property is important for our multi-level message-passing scheme, which we describe next.

\begin{figure}[t]
  \centering
  \includegraphics[width=1.\linewidth]{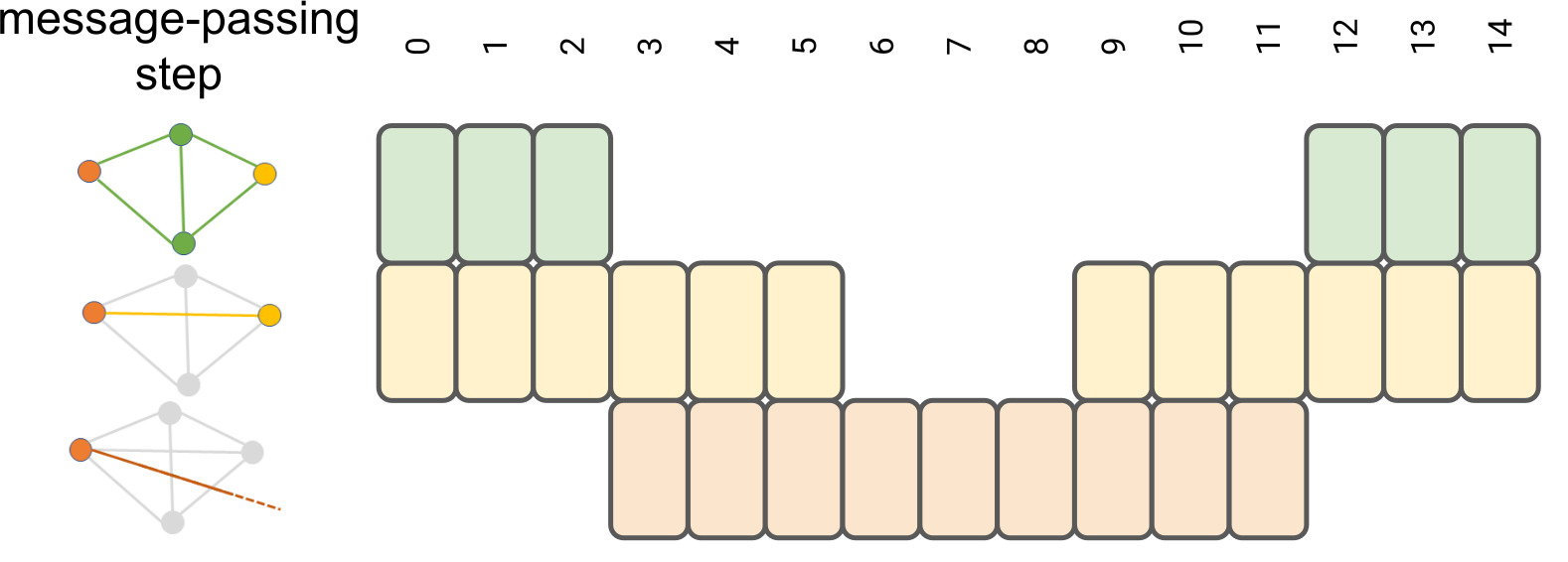}

  \caption{Our hierarchical network architecture with 1 fine (\textit{green}) and 2 coarse (\textit{yellow, orange}) levels. We use 15 message-passing steps, simultaneously processing two levels at a time. 
}
  \label{fig:multiLevelMessagePassing}
\end{figure}

\textbf{Multi-level Message Passing.}
Our nested hierarchical graph representation enables accelerated message-passing through simultaneous processing on multiple levels.
To this end, we endow each level $l$ in the graph with its own set of edge feature vectors $e^l_{ij}$ while the node feature vectors $v_i$ are shared across all levels. 
At the beginning of each message-passing step, we first update edge features on all levels using the finest-level node features: 
\begin{equation}
e^l_{ij}\leftarrow f^l_{v\rightarrow e}(e^l_{ij},v^0_i,v^0_j) \ ,
\end{equation}
where $f^l_{v\rightarrow e}$ is a level-specific MLP. 
Then, node features are updated: 
\begin{equation}
v_{i}\leftarrow f_{e\rightarrow v}(v_{i},\sum_j e^{body}_{ij},\sum_j e^1_{ij},...,\sum_j e^L_{ij}) \ ,
\end{equation}
where $L$ is the number of levels processed in this step. Note that, for each message-passing step, we update the set of body edges $e^{body}_{ij}$ to keep only those that are connected to the currently processed garment nodes.

This scheme has the important advantage of not requiring any explicit averaging or interpolation operators for inter-level transfer. Thanks to the nesting property of our hierarchical graph, nodes are shared across levels and all information transfer happens implicitly by processing the $f_{e\rightarrow v}$ MLPs at the end of each message-passing step.

Our multi-level message-passing scheme can operate on any number of levels simultaneously. We found that the UNet-like architecture shown Fig.~\ref{fig:multiLevelMessagePassing}, with a three-level hierarchy and simultaneous message-passing on two adjacent levels at a time,
yields a favorable trade-off between inference time and quality of results.

To compute the propagation radius of our multi-level message-passing scheme, we sum the maximal distances a message can travel per message-passing step. 
For roughly the same amount of computation,
our architecture yields a radius of 48 edges compared with 15 edges for a single-level scheme.
See Sec.~\ref{sect:experiments} and the supplementary material for a more detailed analysis.

\subsection{Garment Model}
\label{subsec:garmentmodel}
To learn the dynamics of garments, we must model their mechanical behaviour; i.e., the relation between internal deformations and elastic energy, as well as friction and contact with the underlying body. 
Our approach follows standard practice in cloth simulation: we model resistance to stretching with an energy term $\mathcal{L}_{stretching}$, using triangle finite elements \cite{Narain12Adaptive} with a St.~Venant-Kirchhoff material \cite{Montes20Skintight}. 
The $\mathcal{L}_{bending}$ term penalizes changes in discrete curvature as a function of the dihedral angle between edge-adjacent triangles \cite{Grinspun03Discrete}. 
To prevent interpenetration between cloth and body, we penalize the negative distance between garment nodes and their closest point on the body mesh with a cubic energy term $\mathcal{L}_{collision}$, once it falls below a threshold value \cite{santesteban2022snug}. Furthermore, $\mathcal{L}_{inertia}$ is an energy term whose gradient with respect to $\mathbf{x}^{t+1}$ yields inertial forces. Finally, we introduce a friction term $\mathcal{L}_{friction}$ that penalizes the tangential motion of garment nodes over the body \cite{Brown18Accurate,Geilinger20ADD}.

To model the evolution of clothing through time, we follow \cite{Santesteban_2021_CVPR} and use the optimization-based formulation of Martin et al.~\cite{Martin11ExampleBased} to construct an incremental potential
\begin{align}
\label{eq:loss}
\nonumber
   \mathcal{L}_{total} =  & \mathcal{L}_{stretching}(\mathbf{x}^{t+1}) + \mathcal{L}_{bending}(\mathbf{x}^{t+1}) + \\
   & \mathcal{L}_{gravity}(\mathbf{x}^{t+1}) +  \mathcal{L}_{friction}(\mathbf{x}^{t},\mathbf{x}^{t+1}) + \\
   \nonumber
   &\mathcal{L}_{collision}(\mathbf{x}^{t},\mathbf{x}^{t+1}) + \mathcal{L}_{inertia}(\mathbf{x}^{t-1},\mathbf{x}^{t},\mathbf{x}^{t+1})\ ,
\end{align}
where $\mathbf{x}^{t-1}$, $\mathbf{x}^t$, and $\mathbf{x}^{t+1}$ are nodal positions at the previous, current, and next time steps, respectively. 
Minimizing $\mathcal{L}_{total}$ with respect to end-of-step positions is equivalent to solving the implicit Euler update equations, providing a robust method for forward simulation.
When used as a loss during training, this incremental potential allows the network to learn the dynamics of clothing without supervision.

\subsection{Training} 
\label{sect:training}


We train our hierarchical graph network in a fully self-supervised way using the physics-based loss function (\ref{eq:loss}). We briefly discuss some aspects specific to our setting below and provide more details in the supplemental material.

\textbf{Training Data.} 
We use the same set of 52 body pose sequences from the AMASS dataset~\cite{mahmood2019amass} used in \cite{Santesteban_2021_CVPR}. For each training step, we randomly sample SMPL \cite{SMPL:2015} shape parameters $\beta$ from the uniform distribution $\mathcal{U}(-2,2)$. 

We randomly select a garment mesh from a set of templates that includes a shirt, tank top, long-sleeve top, shorts, pants, and a dress. We resize each garment according to the blend shapes of the underlying body and apply a small sizing perturbation randomly sampled from $\mathcal{U}(0.9, 1.1)$ to emulate tighter and looser fits.

We also sample material parameters  (Lam{\'e} parameters for stretching, bending stiffness, and mass density) from log-uniform distributions and use corresponding values in the loss function (\ref{eq:loss}). The same values are attached to all feature vectors, encouraging the network to learn the map between local material parameters and dynamics.

\textbf{Garment Initialization.}
During training, we want to step garment meshes forward in time from any point in the pose sequences. To evaluate the loss function at an arbitrary starting point, we must provide garment geometry for the two previous time steps. We approximate these geometries using linear blend skinning combined with the diffused body model formulation from~\cite{Santesteban_2021_CVPR}. We then remove any intersections between skinned garment meshes and the body.
The initial garment meshes computed in this way are generally not in energetically optimal states. To reduce this internal energy, we find it useful to scale down the contribution of the inertia term using a coefficient $\alpha\in(0,1]$. Using a smaller coefficient $\alpha$ for the first step during training iterations allows the garment to quickly relax into a state of lower potential energy. We also pass $\alpha$ as an input to the network (see Sec.~\ref{subsec:graphnet}) so it can adapt its prediction to different values.
Finally, it should be noted that skinning is only used for initialization during training, but not at inference time.

\textbf{Normalization.} 
Following~\cite{pfaff2020learning}, we find it crucial for convergence to normalize feature vectors at the beginning of each step using exponentially weighted averages of their mean and standard deviations. We also perform denormalization on the outputs to obtain accelerations. Since we do not have access to ground-truth data, we use statistics collected from linearly-skinned garments as described above.

\textbf{Autoregressive Training.}
We start by predicting accelerations for only one next step, then gradually increase the number of predicted steps every $k$ iterations up to 5. We set $k$ to 5000 and find that predicting a maximum of 5 steps during training is enough for the model to autoregressively run for hundreds of steps at inference time.

%% file: 04_experiments.tex
\section{Evaluation}
\label{sect:experiments}

To evaluate HOOD, we first perform a perceptual study that compares the perceived visual quality of our results with the state of the art.
Next, we analyze the benefit of our network architecture using quantitative performance metrics. Finally, we show examples that demonstrate the main advantage of HOOD; i.e., its generalization capabilities.

\subsection{Comparison to State-of-the-Art Methods}

\begin{figure}[t]
  \centerline{  \includegraphics[width=1\linewidth]{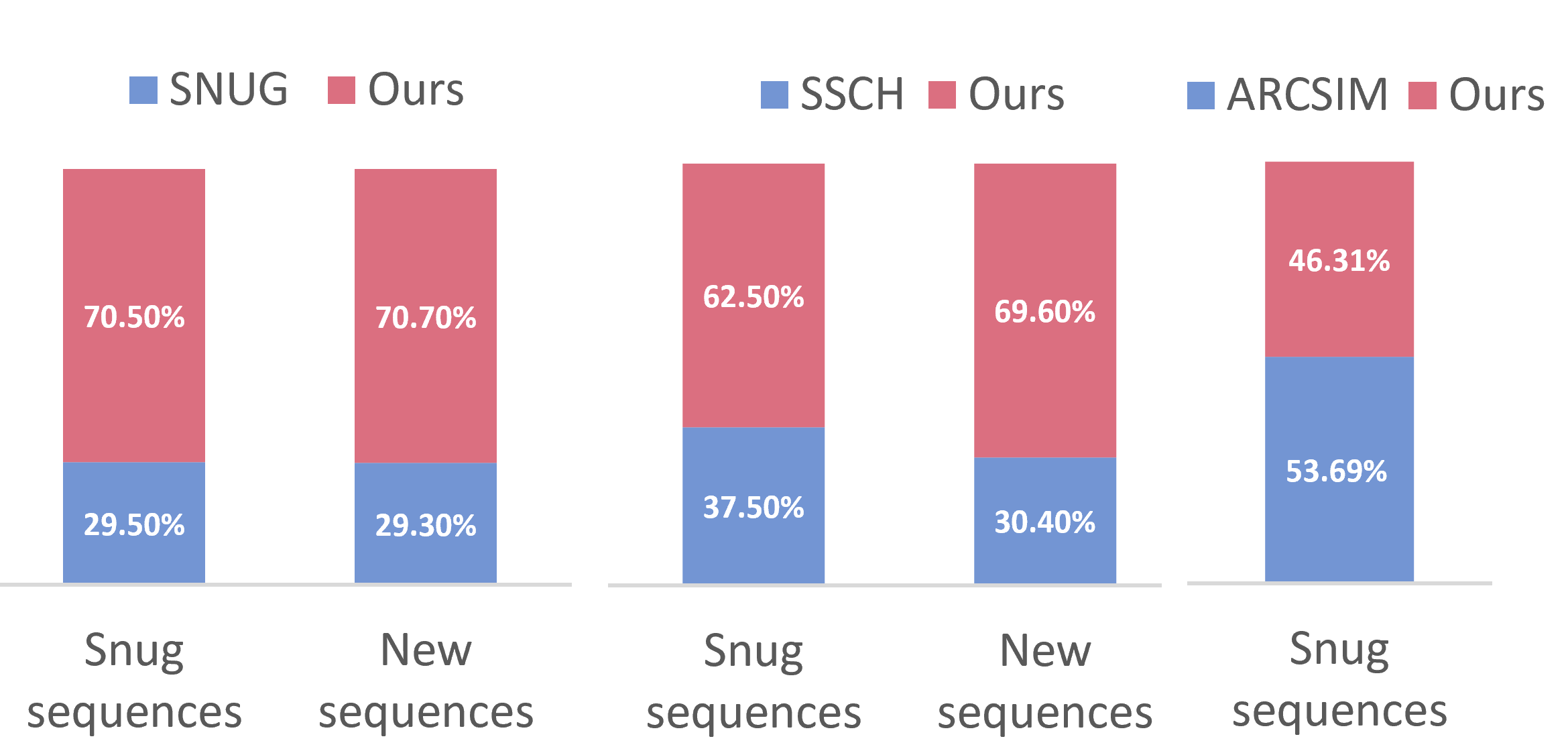}}
\vspace{-0.07in}
  \caption{\textbf{Perceptual study.} 
  Each bar shows the percentage of sequences for which participants preferred our method over the baseline, i.e., SNUG~\cite{santesteban2022snug}, SSCH~\cite{Santesteban_2021_CVPR} or ARCSIM~\cite{narain2012adaptive}. Our method comfortably outperforms learned approaches while being on par with a genuine physical simulator. We do not compare to ARCSIM on the new sequences due to a large number of self-collisions in them, that ARCSIM failed to resolve.}
  \label{fig:userstudy}
\end{figure}

To evaluate the quality of the dynamic garment motions produced by HOOD, we compare it to two state-of-the-art alternatives,  SSCH~\cite{Santesteban_2021_CVPR} and SNUG~\cite{santesteban2022snug}. 
Since there are no object measures of quality and plausibility, we use a perceptual study:
30 participants were shown pairs of videos, generated by our method and a baseline, for the same garment and pose sequence in a side-by-side view. Participants were asked to select the sequence in which \textit{"the garment looks and behaves more realistically"}. The order and placement (left/right) of the videos were randomized per presentation.

We used 8 body pose sequences from the AMASS dataset~\cite{mahmood2019amass} for this study: 4 sequences from the validation set used in \cite{santesteban2022snug}, and 4 new sequences with more dynamic and challenging body motions, including upside-down poses. We analyze preferences from participants separately for the two sequence sets. 

In each of the comparisons, we use the full set of garments that the baseline was trained on. For SNUG, this includes 5 garments (t-shirt, long-sleeve, tank top, pants, shorts), for SSCH it includes 2 garments (t-shirt, dress). For each sequence, we randomly sample SMPL shape parameters $\beta$ from the uniform distribution $\mathcal{U}(-2,2)$.

\begin{figure}[t]
  \centerline{  \includegraphics[width=1\linewidth]{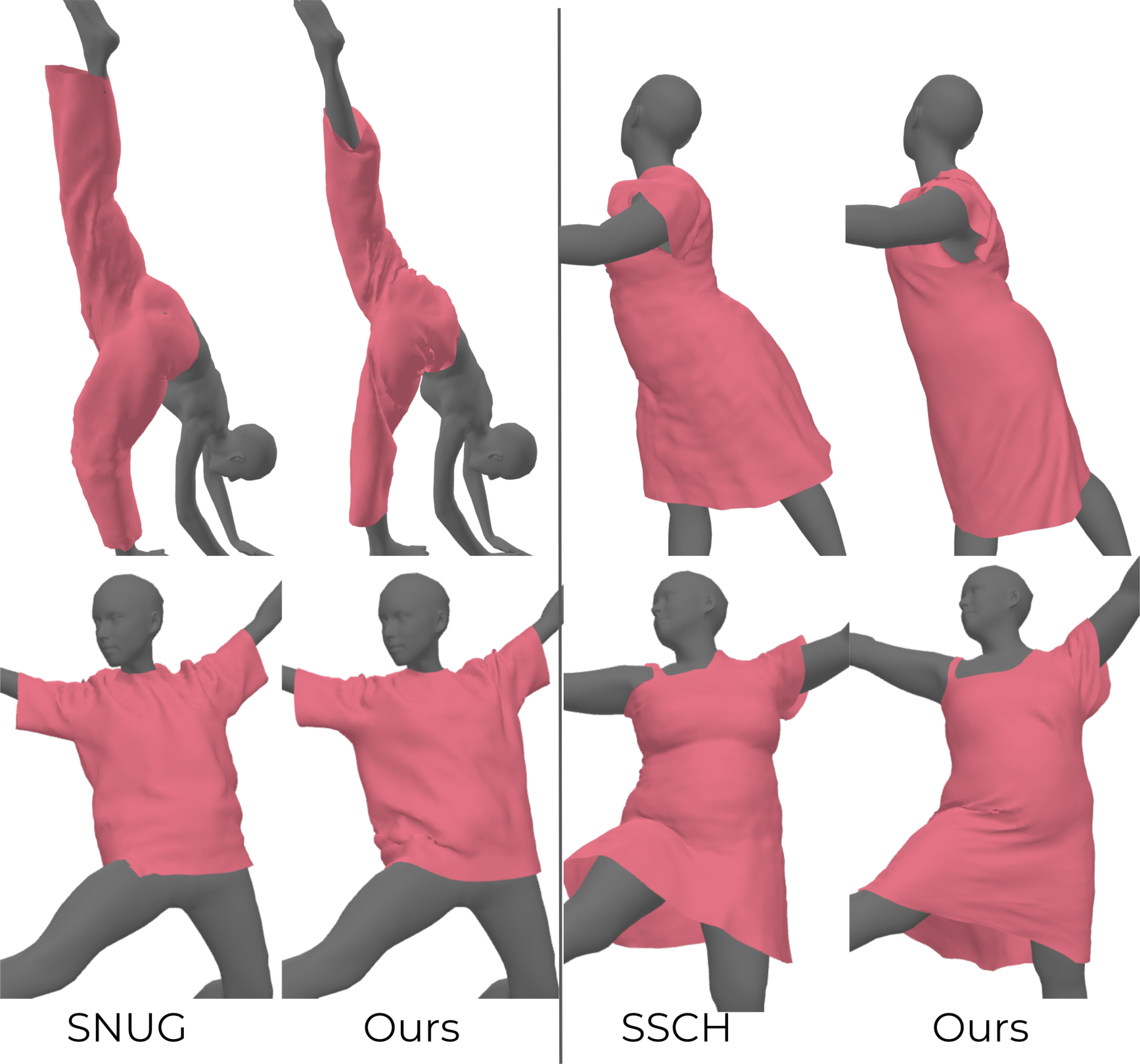}}
\vspace{-0.07in}
  \caption{Qualitative comparison to state-of-the-art methods, SNUG~\cite{santesteban2022snug} and SSCH~\cite{Santesteban_2021_CVPR}. Note how our geometry conforms well to the body and flows naturally with plausible wrinkles in areas without body contact.}
  \label{fig:sotacomp}
\end{figure}

As can be seen from \cref{fig:userstudy}, participants preferred results from our method in a clear majority of the cases when compared to other learned methods, whereas the preferences among our method and a physical simulator ARCSIM~\cite{narain2012adaptive} are nearly equal.
To put this result into perspective, we note that all of our results were generated by a single network, while the learned baselines need a separate trained model for each garment. 
In \cref{fig:sotacomp} we also demonstrate the qualitative advantages of our method over the state-of-the-art.

\begin{table}[]
    \centering
    \resizebox{1\linewidth}{!}{
\begin{tabular}{c|>{\centering}m{0.17\linewidth}|>{\centering}m{0.12\linewidth} >{\centering}m{0.19\linewidth}|>{\centering}m{0.11\linewidth}|cccc}
                        & garments                    & \# steps & propagation radius $\uparrow$& average speed, fps $\uparrow$ & $\mathcal{L}_{total} \downarrow$             & $\|\dfrac{\partial \mathcal{L}_{total}}{\partial \hat{x}}\|_2 \downarrow$       \\ \hline
\textit{Fine15} & \multirow{3}{*}{only dress} & 15       & 15              &         6.65           & 2.92 &     2.37e-2    \\
\textit{Fine48} &                             & 48       & 48              &         2.45           & \textbf{1.43} &  \textbf{1.03e-2}      \\
\textit{Ours}     &                             & 15       & 48              &         \textbf{7.27}  & 1.48  & 1.56e-2 \\ \hline
\textit{Fine15} & \multirow{3}{*}{all}        & 15       & 15              &         13.1           & 1.68  &  3.7e-2     \\
\textit{Fine48} &                             & 48       & 48              &         4.99           & \textbf{1.04} &  \textbf{2.5e-2}       \\
\textit{Ours}     &                             & 15       & 48              &         \textbf{13.6}  & 1.07 & 2.72e-2
\end{tabular}
    }
    \vspace{-0.07in}
    \caption{Comparison of our hierarchical model to two ablations in terms of total loss values and inference speed for a dress with 12K vertices from the VTO dataset~\cite{Santesteban_2021_CVPR}. See supplementary material for more detail.}
    \label{table:coarse}
\end{table}

\begin{figure}[t]
  \centerline{  \includegraphics[width=1.\linewidth]{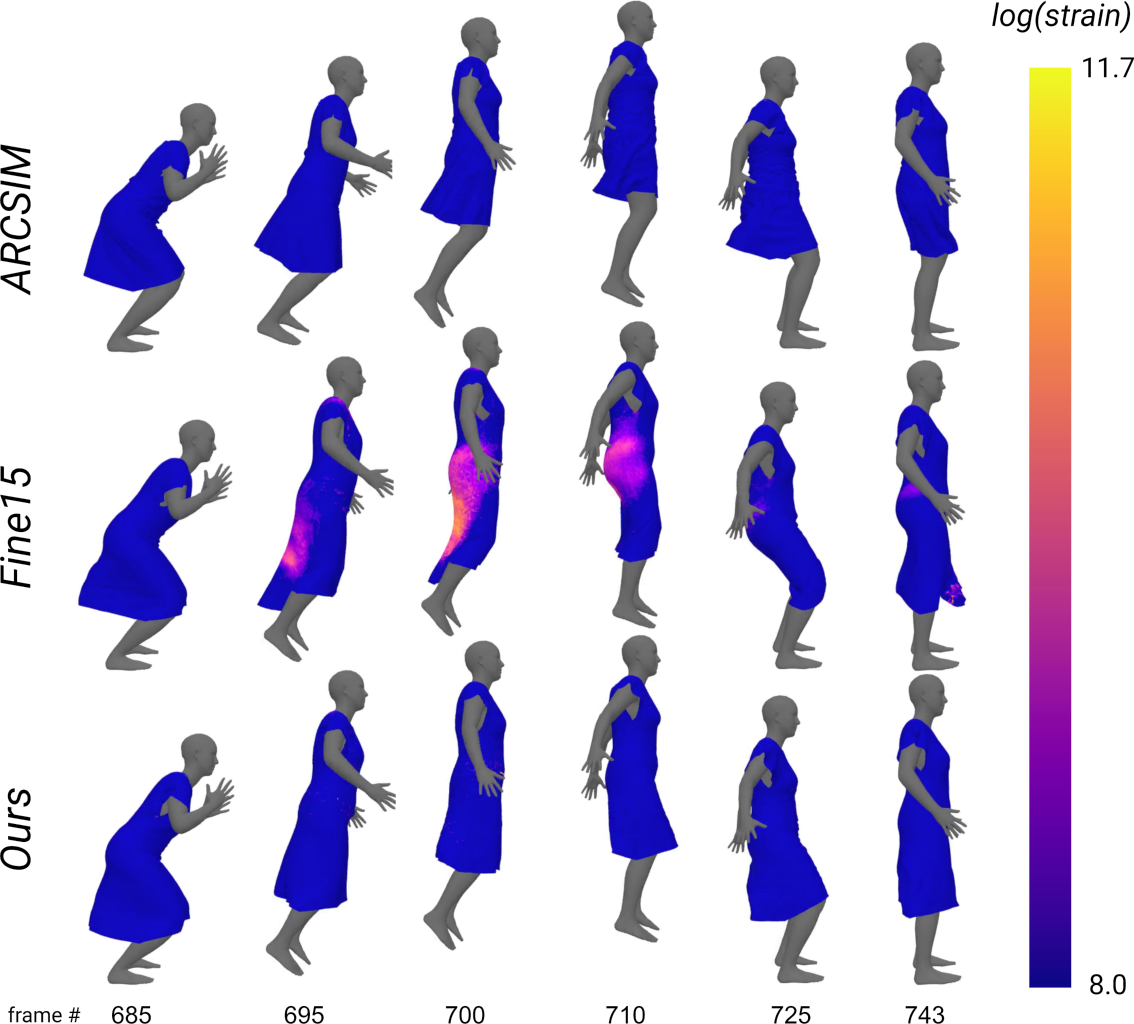}}
\vspace{-0.05in}
  \caption{Qualitative comparison of our hierarchical network to the baseline (\textit{Fine15}) using only a single level. The baseline method requires more time steps to propagate messages from the point where contact between cloth and body occurs to the loose parts of the dress. This results in large stretching artifacts (indicated in pink) and rubbery material behaviour. We also show results of a physical simulator, ARCSIM~\cite{narain2012adaptive}, for reference.}
  \label{fig:strain}
\end{figure}

\subsection{Hierarchical Architecture}
To evaluate our multi-level message passing, we compare our hierarchical UNet-like architecture to two baseline models in terms of objective value and gradient norm, averaged across the validation set.  
The ablations use the single-level message passing scheme from MeshGraphNets~\cite{pfaff2020learning} with 15 (\textit{Fine15}) and 48 (\textit{Fine48}) steps, respectively. 
Table \ref{table:coarse} shows that our hierarchical model outperforms \textit{Fine15} in both inference speed and loss value. 
This is due to the fact that our hierarchical method has a propagation radius of 48 edges per time step, compared to only 15 for the single-level baseline. 
When increasing the number of message passing steps to 48, the single-level baseline \textit{Fine48} achieves slightly better scores for loss and gradient norm, but its inference speed is significantly lower.
\Cref{fig:strain} shows how the limited propagation radius of the baseline \textit{Fine15} results in excessive stretching for sequences with dynamic motions.

\subsection{Generalization}

\begin{figure}[t]
  \centerline{  \includegraphics[width=1\linewidth]{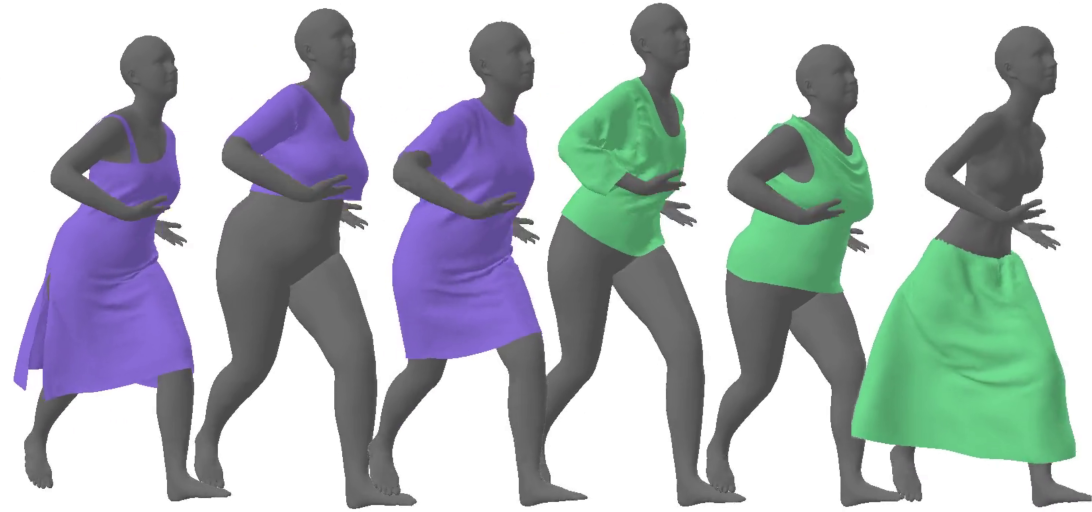}}
\vspace{-0.05in}
  \caption{Results generated by HOOD for garments unseen during training. Garments in purple are modified versions of those from the training set, while garments in green are completely new.}
  \label{fig:generalization}
\end{figure}

\begin{figure}[t]
  \centerline{  \includegraphics[width=1\linewidth]{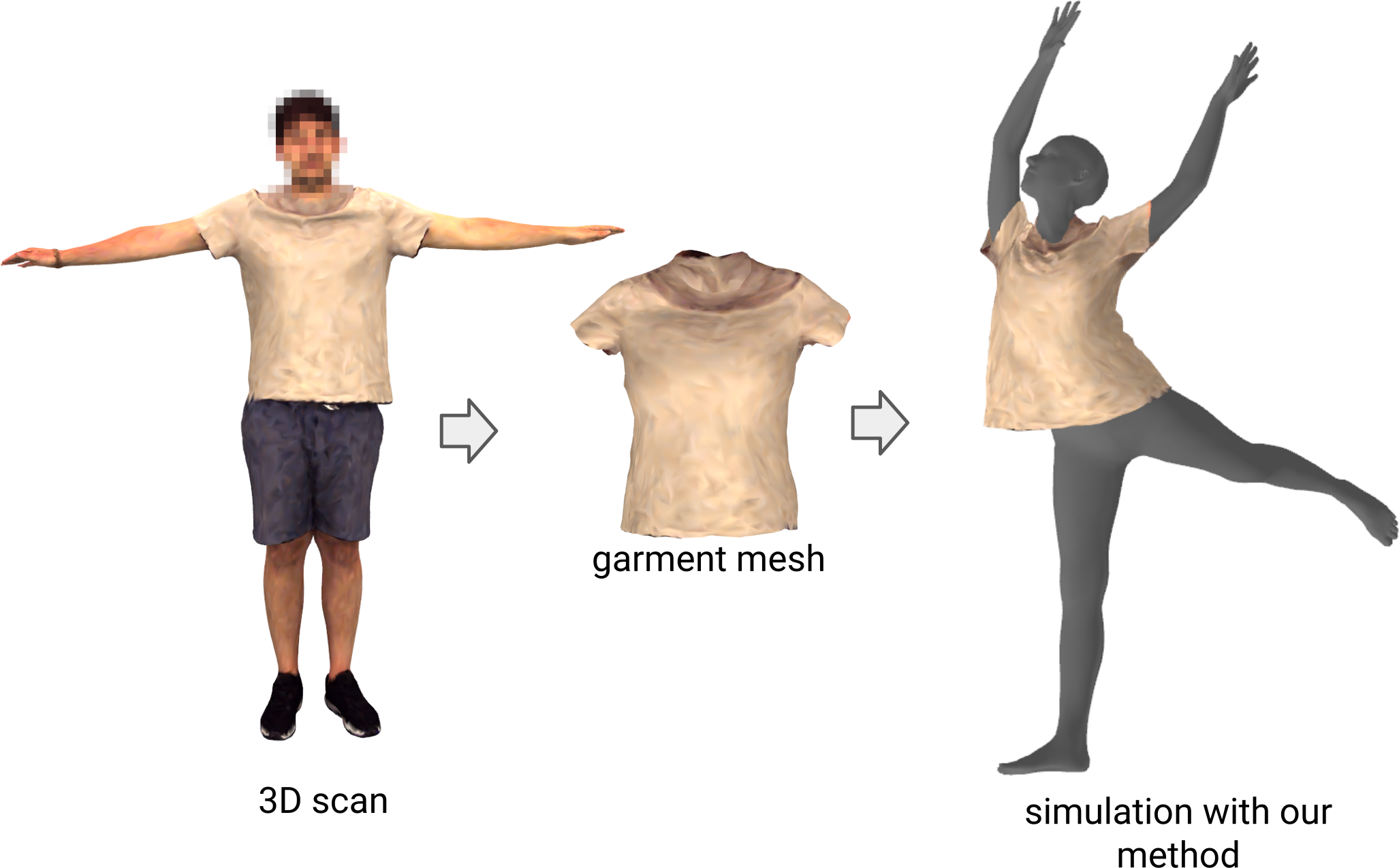}}
  \vspace{-0.05in}
  \caption{We manually extract a garment mesh from a 3D scan and animate it with HOOD.}
  \label{fig:scan}
\end{figure}

\noindent\textbf{Novel Garments.}
Since our method learns the local behaviour of fabric, it is not limited to inferring motion for the garments from the training set. 
\Cref{fig:generalization} shows results generated by HOOD for garments not seen during training. The three garments on the left are \emph{modified} from the training set (\textit{dress} with a cut, \textit{long sleeve top} with shorter sleeves, and elongated \textit{t-shirt}). The next two are provided by the authors of FoldSketch~\cite{Li2018FoldSketch}. The rightmost skirt was manually created in Blender.
HOOD can also simulate real-world garments; 
e.g.~Figure~\ref{fig:scan} shows the animation of a garment manually extracted from a 3D scan.

\noindent\textbf{Resizing Garments.}
Since our model takes the garment's edge lengths at rest as input, it is possible to resize the rest mesh at inference time to simulate different sizes of the same garment. We provide examples in the supplementary material and the accompanying video.

\noindent\textbf{Changing Topology.}
We represent garments as a graph, therefore we can naturally handle dynamically changing topology (e.g., through buttons or zippers). 
For example, in \cref{fig:unzip}, we  manually choose pairs of vertices to connect graph segments (the ``buttons''). Toggling inclusion of these edges ``buttons'' or ``unbuttons'' the garment.

\begin{figure}[t]
  \centerline{  \includegraphics[width=1\linewidth]{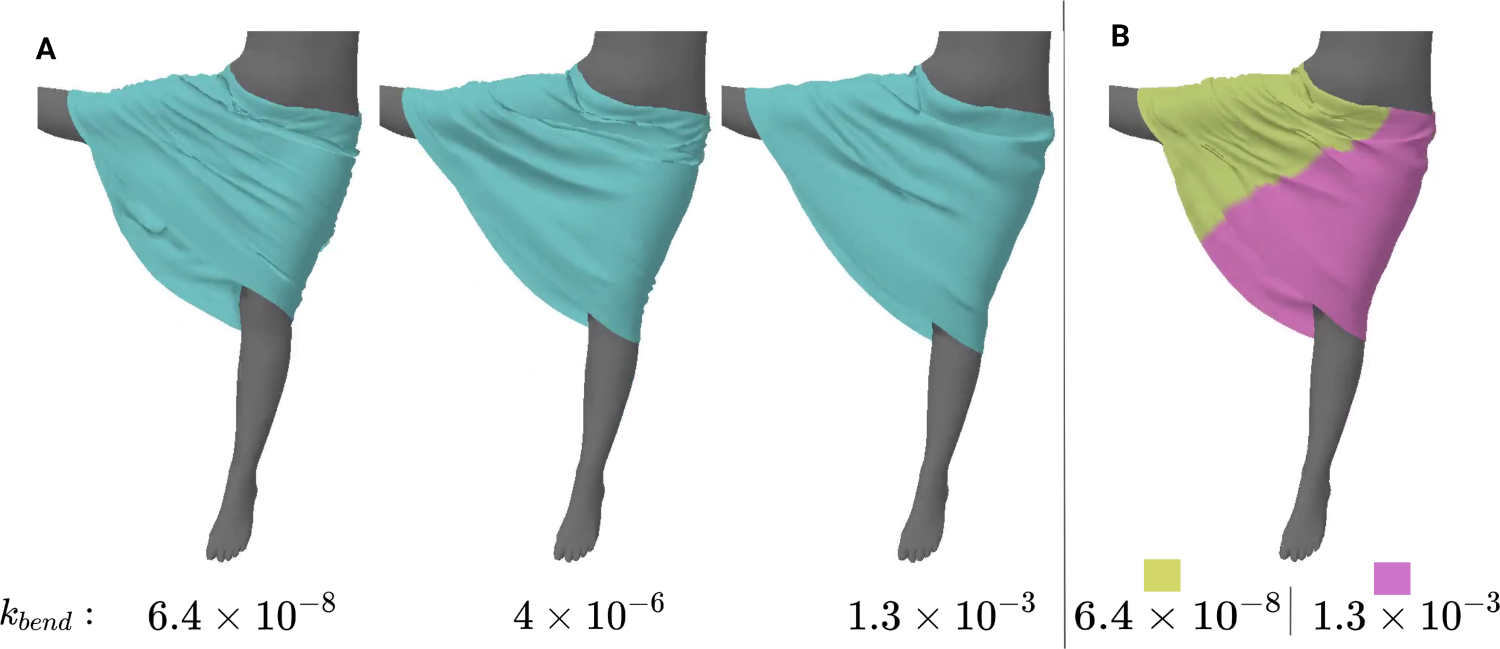}}
\vspace{-0.05in}
  \caption{\textbf{A}: A single network can model garments with different material parameters. \textbf{B}: The network is also able to model garments made of several materials. Here two halves of the dress (in yellow and purple) have different bending coefficients.}
  \label{fig:bcoef}
\end{figure}

\noindent\textbf{Material parameters.}
By augmenting the input material parameters and applying corresponding supervision during training, HOOD can simulate garments with different materials. Furthermore, since the input material parameters are defined separately for each vertex and edge in the garment mesh, we can define different materials for different parts of a single garment.
\cref{fig:bcoef} (A) shows an example where our method---a single trained network---produces qualitatively different folds and wrinkle patterns when given different bending coefficients. 
We also show an example of a single garment with different materials for different parts in~\cref{fig:bcoef} (B).

%% file: 05_limitations.tex
\section{Conclusion and Limitations}
\label{sect:limitations}
We have proposed a novel method for modeling the dynamics of virtual garments. While previous methods have  shown the benefits of combining physical simulation and machine learning, they often do not generalize to unseen garments and struggle with loose and free-flowing garments due to their reliance on skinning.  
To overcome these limitations, we leverage a graph-based garment representation that is processed by a graph neural network to predict nodal accelerations. We further introduce a hierarchical graph representation that accelerates the propagation of stiff waves through the garment and, consequently, leads to more natural simulation results. Following recent work~\cite{santesteban2022snug}, we leverage an optimization-based formulation~\cite{Martin11ExampleBased} of physical simulation for supervision. 
Together, these contributions enable HOOD to predict plausible dynamic motion for a wide range of garments on arbitrary body shapes, while generalizing to unseen garments and new pose sequences. 
Our method can also handle dynamic changes in material properties and garment topology at inference time. We believe that HOOD constitutes an important step towards the goal of modeling realistic and compelling clothing behavior on digital humans.

\smallskip
\noindent\textbf{Limitations.} Our method currently has a number of limitations that we plan to address in future work.
First, our model does not handle garment-garment interactions and our results may therefore exhibit self-penetrations. While MeshGraphNets~\cite{pfaff2020learning} can model cloth self-collisions, the method learns this behavior from ground-truth data generated from an offline simulation method. 
Since our model is trained in a fully self-supervised way, interaction patterns between remote garment nodes are not known in advance. Na{\"i}vely including large sets of \textit{collision edges} to handle potentially colliding garment nodes would be prohibitively expensive.
Devising methods to deal with garment self-collisions in a self-supervised fashion is an exciting direction for future work.
Second, while our model learns the effects of body-garment interactions well, it may fail when body motions exceed the velocities seen at training time. One promising strategy to address this would be to introduce continuous collision detection into our approach.
%
Finally, when simulating garments on bodies with severe self-intersections, our approach may use the wrong body-garment correspondences and produce erroneous results. While the automatic resolution of body self-penetrations is an interesting problem even for physics-based simulation, it is beyond the scope of this work. 

\noindent\textbf{Acknowledgements}  Artur Grigorev was supported by the Max
Planck ETH Center for Learning Systems. 
We thank Xi Wang, Christoph Gebhardt, Velko Vechev, Yao Feng, and Tobias Pfaff for their feedback and help during the project.

\noindent\textbf{Disclosure} MJB has received research gift funds from
Adobe, Intel, Nvidia, Meta/Facebook, and Amazon. MJB
has financial interests in Amazon, Datagen Technologies,
and Meshcapade GmbH. MJB’s research was performed
solely at, and funded solely by, the Max Planck.